\documentclass[%
 preprint,
 amsmath,amssymb,
]{revtex4-1}

\usepackage{graphicx}
\usepackage{amssymb}
\usepackage{setspace}
\usepackage[utf8]{inputenc}
\usepackage{array}
\usepackage{natbib}
\usepackage{amsmath}
\usepackage{multirow}
\usepackage{hyperref}

\usepackage{color}
\definecolor{redcolor}{rgb}{1.0,0.,0.}

\begin{document}

\preprint{}

\title{Language Networks: a Practical Approach}

\author{Jorge A. V. Tohalino$^1$}%
\author{Diego R. Amancio$^1$}

\affiliation{$^1$Institute of Mathematics and Computer Science, University of S\~ao Paulo, S\~ao Carlos, Brazil\\
%
%
}

\date{\today}

\begin{abstract}
This manuscript provides a short and practical introduction to the topic of language networks. This text aims at assisting researchers with no practical experience in text and/or network analysis. We provide a practical tutorial on how to model and characterize texts using network-based features. In this tutorial, we also include examples of pre-processing and network representations.
A brief description of the main tasks allying network science and text analysis is also provided. A further development of this text shall include a practical description of network classification via machine learning methods.
\end{abstract}

\maketitle



\section{Language networks: applications} \label{sec:app}

Complex networks have been used to model a wide range of applications~\cite{battiston2020networks}. Particularly, this model have been used also to analyze and classify texts. From the theoretical perspective, complex networks have been used to analyze many aspects of human language, including the emergence of Zipf's Law~\cite{ferrer2020optimal} and semantic and cognitive properties related to language. Many of such analysis are in the Physics field. Practical approaches includes many applications in Computer Science. More recently, many applications of language network analysis. For reference purpose, we list below some of the applications of language network analysis. We include both theoretical and perspectives approaches. A more detailed account of network-based models can be found e.g. in~\cite{cong2014approaching}.

\begin{enumerate}

    \item \emph{Authorship attribution and style attribution}: this task aims at identifying the authorship of a text, given a set of possible authors in the supervised case. Examples of network-based works addressing this topic include refs. ~\cite{segarra2015authorship,de2016using,marinho2016authorship,akimushkin2018role,machicao2018authorship,akimushkin2017text,amancio2015authorship,segarra2016attributing,segarra2013authorship}.

    \item \emph{Stylometry}: this taks aims at characterizing different styles observed in literary movements, scientific journals and other issues. This topic is also closely related to the authorship attribution task, since authors have different writing style. Stylometric features can also be used e.g. to discriminate real from fake texts. Examples of network-based works addressing this topic include refs. ~\cite{eisen2018stylometric,marinho2018labelled,amancio2015complex,amancio2015probing,amancio2012structure,amancio2015comparing,jiang2019does,stanisz2019linguistic, roxas2012characterizing, stevanak2010distinguishing, roxas2010prose}.

    \item \emph{Complexity analysis}: this task aims at analyzing the features responsible for making a text complex. Research on this topic includes automatic essay scoring and comparison of language complexity patterns. Examples of network-based works addressing this topic includes~\cite{liu2010language,liu2008complexity,amancio2012complex,liu2011can,antiqueira2007strong, cardenas2016does}.

    \item \emph{Document Summarization}: in this task one aims at identifying the most important information from large documents. In recent years complex networks have been used to identify the relevance of words, sentences or paragraphs via network centrality measurements. Examples of network-based works addressing this topic includes~\cite{tohalino2018extractive, tohalino2017extractive, antiqueira2009complex, mihalcea2004graph, amancio2012extractive,mehler2008structural}.

    \item \emph{Language theory}: this line of research includes the study of many aspects of linguistic theory, including the emergence of linguistic and complex systems patterns. Examples of network-based works addressing this topic includes~\cite{stella2015patterns,stella2019forma,stella2019modelling, bochkarev2018modelling,cancho2001small,i2003least,kello2010scaling,i2004euclidean,ferrer2006syntactic,ferrer2007spectral,i2005structure,choudhury2009structure,sole2010language,soares2005network,peng2008networks,li2007chinese,li2005structural,zhou2008empirical,brede2008patterns,sheng2009english,grabska2012complex,liang2012study,gao2014comparison,vcech2009word,wenwenextracting,abramov2011automatic,ke2008analysing,gibson1998linguistic,liu2010dependency,liu2010language}.

    \item \emph{Semantic analysis}: complex networks have also been used to analyze the semantic aspects of texts. This is possible because different models can capture different language aspects~\cite{amancio2012consistencia,correa2019word,correa2020semantic,correa2018word,liu2009statistical,ferrer2001two,borge2010semantic,mihalcea2011graph,sole2015ambiguity,silva2013discriminating}.

    \item \emph{Story flow}: this line of studies how stories unfolds via network analysis. Examples of network-based works addressing this topic includes~\cite{amancio2016network, dekker2018evaluating, holanda2019character, prado2016temporal}. An example of story flow visualization is provided in Figure \ref{fig:meso}.

    \item \emph{Language and cognition}: many cognitive aspects can be studied via language. This includes studies the analysis the relationship between language features and cognitive impairment. Examples of studies on this topic can be found in refs.~\cite{dos2017enriching,stella2019viability,stella2018cohort,stella2020forma, choudhury2007difficult,baronchelli2013networks,sigman2002global,de2004thesaurus,steyvers2005large,gravino2012complex,10.7717/peerj-cs.295, sole2010language}.

    \item \emph{Keyword extraction}: this task is one of the most important and basic tasks in information retrieval. The identification of relevant words can also be important as a sub-task in other tasks, such as in document summarization and document clustering. Examples of studies on this topic can be found in refs.~\cite{lahiri6571keyword,zhan2012keyword,beliga2016selectivity,wang2015using,amancio2013probing,amancio2012extractive,li2016evolutionary,sivsovictoward,yan2016chinese,garg2018identifying,mathiesen2014statistics}.

    \item \emph{Sentiment analysis}: this task uses natural language processing and related textual approach to identify, in a systematic way, affective states in particular words, sentences or larger chunks of texts~\cite{mitrovic2010networks,fornacciari2015social}.

\end{enumerate}

\section{A practical example: Modeling and characterizing co-occurrence networks}

\begin{figure}[h]
    \centering
    \includegraphics[width=0.9\textwidth]{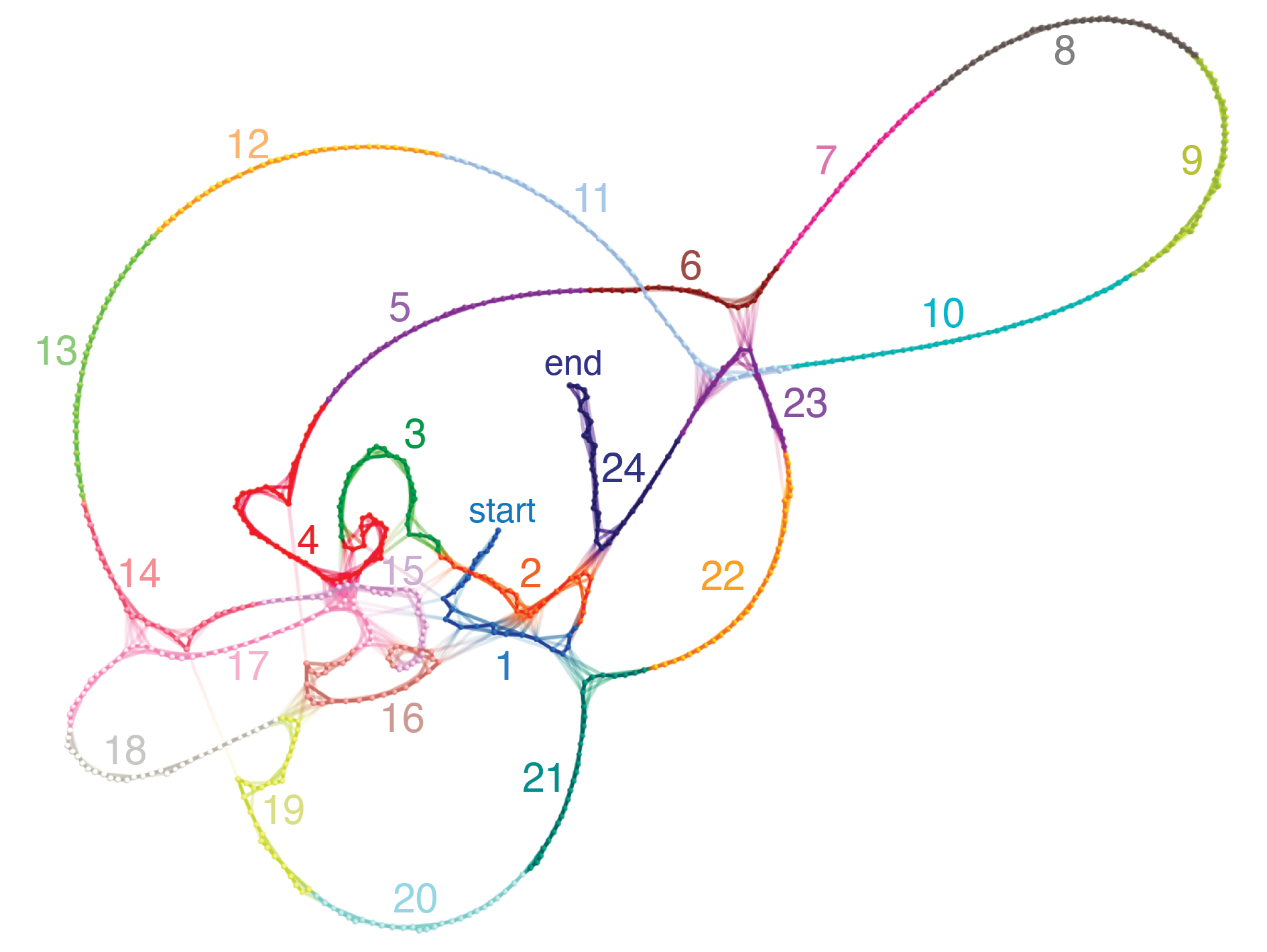}
    \caption{Example of mesoscopic network, where nodes represent a set of consecutive paragraphs and edges are established between semantically similar nodes. The network was obtained
from The Odyssey book series (attributed to Homer). Nodes colors and respective numbers indicate the books they belong to. }
    \label{fig:meso}
\end{figure}

A simple approach to model a text as a complex networks consists in linking adjacent words whenever they appear adjacent in the text.
It has been shown that this type of representation captures mostly stylistic features~\cite{amancio2013probing}. While other types of representation are able of capturing semantic features (see Figure \ref{fig:meso}), our focus here is on co-occurrence networks.
Word adjacency networks can be viewed as a simplification of text networks created via syntactical links, where nodes are words and edges are syntactical dependencies~\cite{i2004patterns}. Because most of syntactical links are between adjacent words, a co-occurrence network can be viewed as a simplification of the syntactical model~\cite{i2004patterns}.

The basic steps in the creation of a word adjacency network include the removal of stopwords and lemmatization. The removal of stopwords in most applications because stopwords may largely affect the structure of networks. In most applications, because they mostly link content words, they can be replaced by edges. However, in some applications, stopwords may play a important role in detecting stylometric features~\cite{segarra2013authorship}. After stopwords are removed, the remaining words are lemmatized so that verbs and nouns are mapped into their infinitive and singular forms, respectively.

In the first subsection of this document we describe how stopwords are removed. The lemmatization process is also detailed in the respective section. We then show how the pre-processed text is mapped into a co-occurrence network.






\subsection{Pre-processing}

Natural Language Toolkit (NLTK) is a well-known text processing Python tool~\cite{loper2002nltk}. In this example we use NLTK to import a list of English stopwords (function words). Note that this is a pre-selected list of words. Some strategies to define stopwords in any sequence of symbols can be also used for this purpose~\cite{amancio2013probing,carretero2013improving}. As first step, we import libraries to handle stopwords and to perform the lemmatization process:

\begin{figure}[h]
    \centering
    \includegraphics[width=0.6\textwidth]{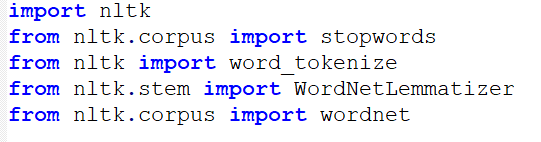}
    \caption{Libraries required to perform the removal of stopwords and the lemmatization process. To split the text into words we use the word tokenize library.}
    \label{fig:ls1}
\end{figure}

The first step in processing the text consists in tokenizing the text so that a list of words is provided as output:

\begin{figure}[h]
    \centering
    \includegraphics[width=0.65\textwidth]{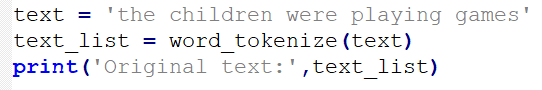}
    \caption{The input text "the children were playing games" is tokenized to form a list of words using the function \emph{word\_tokenize}.}
    \label{fig:ls2}
\end{figure}

\begin{figure}[h]
    \centering
    \includegraphics[width=0.95\textwidth]{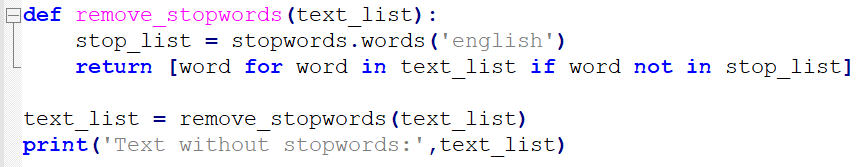}
    \caption{Removing stopwords from the text.}
    \label{fig:ls3}
\end{figure}

The output confirms that the text becomes a list of tokens: ``the'', ``children'', ``were'', ``playing'' and ``games''. In the next step, tokens identified as stopwords are removed from the text (see Figure \ref{fig:ls3}). The output of Figure \ref{fig:ls3} lists the words ``children'', ``playing'' and ``games'', which confirms that ``were'' and ``the'' were identified as stopwords and removed. Note that the original word order is kept even after stopwords are disregarded. This is a essential feature in the construction and analysis of co-occurrence networks.



\begin{figure}[h]
    \centering
    \includegraphics[width=1\textwidth]{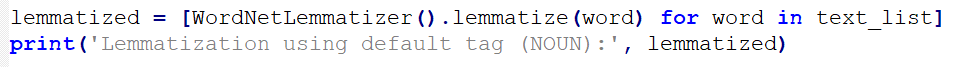}
    \caption{Example of word lemmatization using \textit{WordNetLemmatizer}. The input parameters are the word and its corresponding part-of-speech. Because we did not provide the part-of-speech for each word, all words were regarded as noun.}
    \label{fig:ls4}
\end{figure}

After stopwords are removed, \emph{WordNetLemmatizer} in Figure \ref{fig:ls4} is used to extract the lemma of the remaining words. Here we do not provide the part-of-speech for each word, therefore all words are regarded as noun. The output of the code in Figure \ref{fig:ls4} lists the words ``child'', ``playing'' and ``game''.

\begin{figure}[h]
    \centering
    \includegraphics[width=0.75\textwidth]{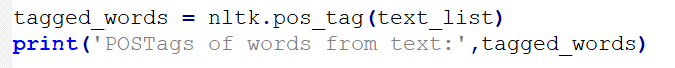}
    \caption{Example of part-of-speech tags obtained for the considered input words.}
    \label{fig:ls_21_22}
\end{figure}

The output reveals that the lemmatization process is not conducted with success for all words. Note that the word ``playing'' is not lemmatized since its part-of-speech is incorrect. In order to provide the correct part-of-speech for each word, we use \emph{POSTagging} from NLTK, as shown in Figure \ref{fig:ls_21_22}. The output lists the part-of-speech of each word in a list format, i.e. ('children', 'NNS'), ('playing', 'VBG') and ('games', 'NNS').

\begin{figure}[h]
    \centering
    \includegraphics[width=0.5\textwidth]{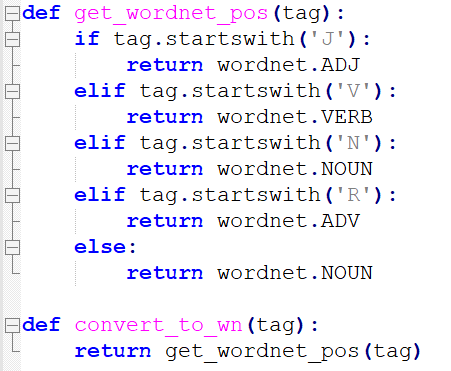}
    \caption{Functions used for generating ``N'', ``V'', ``J'' and ``R'' respectively for nouns, verbs, adjectives and adverbs.}
    \label{fig:ls_24_37}
\end{figure}

\begin{figure}[h]
    \centering
    \includegraphics[width=0.94\textwidth]{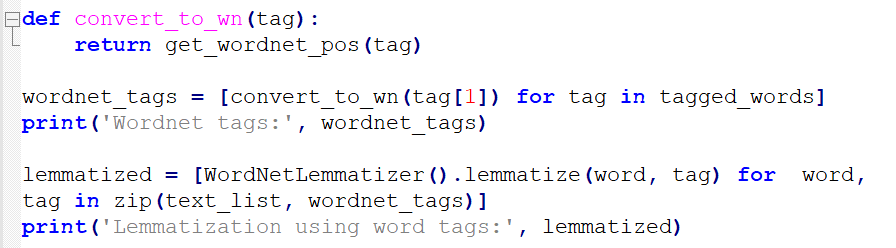}
    \caption{Converting part-of-speech tags and performing the lemmatization process.}
    \label{fig:ls_39_40}
\end{figure}

In order to use the obtained tags in \textit{WordNetLemmatizer}, the following tags should be provided: ``N'', ``V'', ``J'' and ``R'' respectively for nouns, verbs, adjectives and adverbs. To map the output of the function used for pos-tagging into ``N'', ``V'', ``J'' and ``R'', we can use Figures \ref{fig:ls_24_37} and \ref{fig:ls_39_40}. In particular, Figure \ref{fig:ls_39_40} uses the code in Figure \ref{fig:ls_24_37} to convert part-of-speech tags generated by NLTK to the WordNet format. The ouptut of Figure \ref{fig:ls_39_40} provides the following list of tags: 'n', 'v' and 'n'. Finally, once the obtained tags are mapped to the desired format, they can be used to perform the lemmatization process using Figure \ref{fig:ls_39_40}. Now the output lists the words ``child'', ``play'' and ``game'' (in this order), as expected.

\begin{figure}[h]
    \centering
    \includegraphics[width=0.95\textwidth]{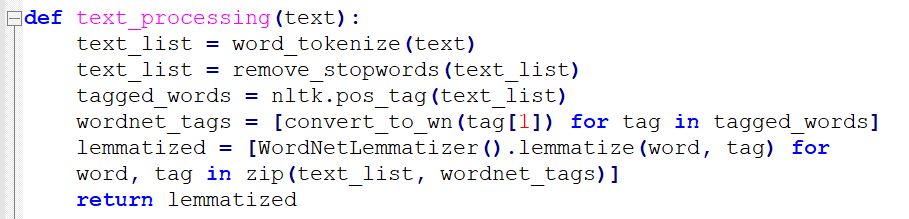}
    \caption{Function used to pre-process a text. The pre-processing comprises the steps of (i) removing stopwords; (ii) part-of-speech tagging; and (iii) lemmatization. In some network models, the removal of stopwords is an optional step.}
    \label{fig:ls_45_51}
\end{figure}

We summarize in Figure \ref{fig:ls_45_51} all the steps required to pre-process a text before it can be used to create co-occurrence networks. The steps include (i)  removal of stopwords (this is optional, depending on the model being used), (ii) part-of-speech tagging, and (iii) lemmatization.

\subsection{Creating a co-occurrence text-network}

Co-occurrence networks are well-known network models used to represent texts as complex networks. This model represents nodes as words and edges are established between adjacent (or nearby) words. Some works use windows so that words in the same window are linked as a clique. In the previous sentence, for example, the three first words would generate the following edges: ``some'' -- ``works'', ``works'' -- ``use'' and ``some'' -- ``use'' (we disregarded the pre-processing steps in this example). Usually, the window size ($w$) ranges between 1 and 3.

\begin{figure}[h]
    \centering
    \includegraphics[width=0.65\textwidth]{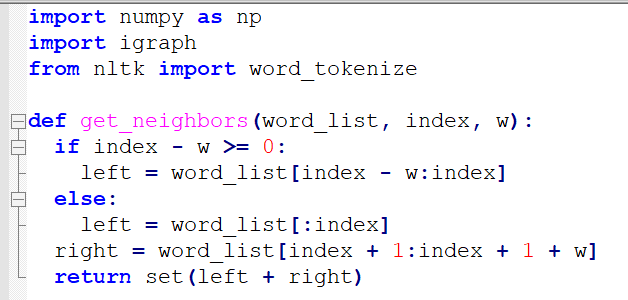}
    \caption{This function take as input parameters a list of words (\textit{word\_list}), the index of the target word in the list (\textit{index}), and the window size $w$.}
    \label{fig:ls_1_11}
\end{figure}

\begin{figure}[h]
    \centering
    \includegraphics[width=1\textwidth]{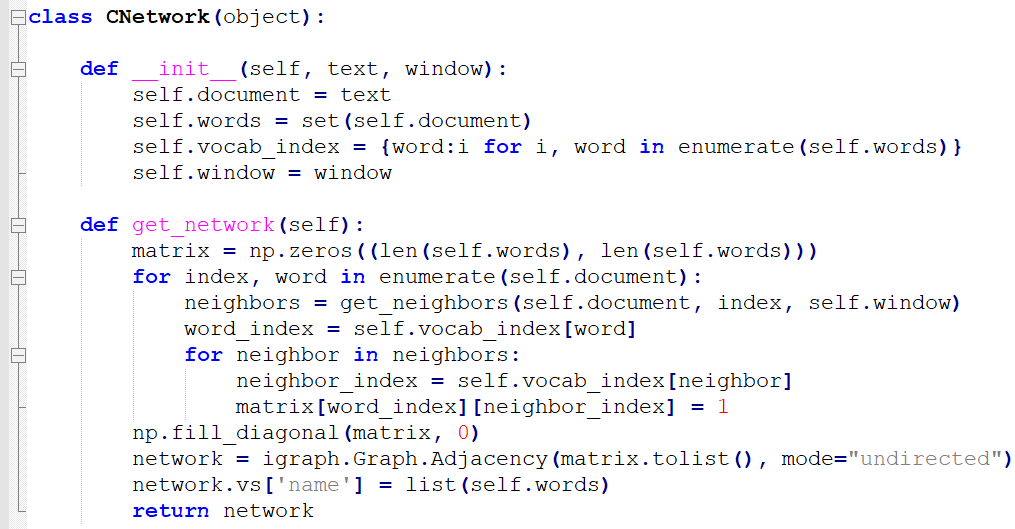}
    \caption{\textit{CNetwork} creates a network model from a raw text.   The function returns the obtained network,which  is  represented  using  igraph  library.}
    \label{fig:ls_13_32}
\end{figure}

Before implementing a function to create a network from a text, we create an auxiliary function \textit{get\_neighbors}. It returns the $w$ nearest neighbors of a word in a word list. This function takes into account both left and right neighbors. This implementation is provided in Figure \ref{fig:ls_1_11}.
\textit{CNetwork} -- the class responsible for mapping a text into a network -- has as inputs the raw text and the window size $w$. The function returns the obtained network, which is represented using igraph library~\cite{csardi2006igraph} (see Figure \ref{fig:ls_13_32}).
\textit{get\_network()} creates a adjacency matrix in order to store adjacency information. Then the function seeks the $w$ nearest neighbors for each word. This is used to define the neighbor nodes for each word.
The obtained matrix is used as input to the function \textit{igraph.Graph.Adjacency()}, which returns the network representation in the igraph format. 

\begin{figure}[h]
    \centering
    \includegraphics[width=1\textwidth]{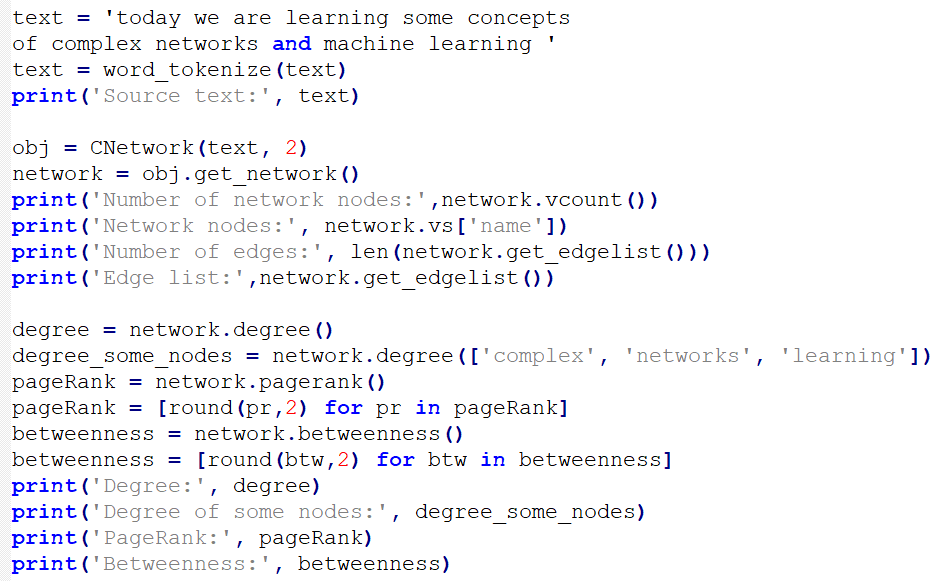}
    \caption{Example of network measurements extracted from the co-occurrence text network.}
    \label{fig:ls_38_43}
\end{figure}

The application of \textit{CNetwork} is illustrated in Figure \ref{fig:ls_38_43}. We start by creating a new text that shall be mapped into a co-occurrence network. The tokenization process in Figure \ref{fig:ls_38_43} provides the following list of words: 'today', 'we', 'are', 'learning', 'some', 'concepts', 'of', 'complex', 'networks', 'and', 'machine' and 'learning'.
The tokenized text is then used as input to \textit{cNetwork}. In this example, we use $w=2$. The following result is obtained after modeling the text as a network:

\begin{enumerate}
    \item Number of network nodes: 11.

    \item Network nodes: 'of', 'networks', 'concepts', 'complex', 'and', 'machine', 'learning', 'some', 'are', 'we' and 'today'.

    \item Number of edges: 21.

    \item Edge list: (0, 1), (0, 2), (0, 3), (0, 7), (1, 3), (1, 4), (1, 5), (2, 3), (2, 6), (2, 7), (3, 4), (4, 5), (4, 6), (5, 6), (6, 7), (6, 8), (6, 9), (7, 8), (8, 9), (8, 10) and (9, 10).

\end{enumerate}

%


In Figure \ref{fig:ls_38_43}, we also list some network measurements extracted from the obtained co-occurrence network. A detailed description of network measurements can be found in ref.~\cite{costa2007characterization}. The obtained values for the considered network measurements are:

\begin{enumerate}

\item Degree: 4, 4, 4, 3, 4, 3, 6, 4, 4, 2, 4.

\item Degree of some nodes: 4, 4, 6.

\item PageRank: 0.09, 0.1, 0.09, 0.07, 0.09, 0.08, 0.14, 0.09, 0.09, 0.06, 0.09.

\item Betweenness: 3.82, 6.27, 1.5, 1.73, 2.0, 2.57, 17.47, 5.37, 2.95, 0.0, 3.33.

\end{enumerate}



\section{Final Remarks}

This text is intended to be a didactic approach for those interested in working at the intersection of network science, natural language processing, pattern recognition and applications. A further development of this manuscript will include a practical description of network characterization and classification. This further development shall touch, therefore, many of the applications described in Section \ref{sec:app}. We also intend to include additional models of network representation using advanced recent neural networks embeddings (see e.g. refs.~\cite{correa2020semantic,quispe2020using}).

\section*{Acknowledgments}
We thank F.N. Silva (Indiana University Network Science Institute) for providing Fig. \ref{fig:meso}.

\newpage

\bibliographystyle{ieeetr}







\end{document}